\pdfoutput=1

\documentclass[11pt]{article}

\usepackage[]{emnlp2021}

\usepackage{times}
\usepackage{latexsym}
\usepackage{adjustbox}

\usepackage[T1]{fontenc}

\usepackage[utf8]{inputenc}

\usepackage{microtype}
\usepackage{subcaption}
\usepackage{graphicx}
\usepackage{amsfonts}
\usepackage{amsmath,amsthm}
\usepackage{latexsym}
\usepackage{adjustbox}
\usepackage[normalem]{ulem}

\title{A Neural Network-Based Linguistic Similarity Measure for Entrainment in Conversations}

\author{Mingzhi Yu \and Diane Litman \\
 University of Pittsburgh \\
  \texttt{miy39@pitt.edu} \\ 
  \texttt{dlitman@pitt.edu} \\ \And
  Shuang Ma \and Jian Wu \\
  Microsoft Corporation, Redmond\\
  \texttt{shuama@microsoft.com}\\
  \texttt{jianwu@microsoft.com} 
}

\begin{document}
\maketitle
\begin{abstract}
Linguistic entrainment is a phenomenon where people tend to mimic each other in conversation. The core instrument to quantify entrainment is a linguistic similarity measure between conversational partners. Most of the current similarity measures are based on bag-of-words approaches that rely on linguistic markers, ignoring the overall language structure and dialogue context. To address this issue, we propose to use a neural network model to perform the similarity measure for entrainment. Our model is context-aware, and it further leverages a novel component to learn the shared high-level linguistic features across dialogues. We first investigate the effectiveness of our novel component. Then we use the model to perform similarity measure in a corpus-based entrainment analysis. We observe promising results for both evaluation tasks.
\end{abstract}

\section{Introduction}
Linguistic entrainment is a phenomenon where individuals unconsciously mimic each other in conversation. 
There has been a large amount of research that studies this phenomenon in a wide range of linguistic dimensions such as acoustic and prosodic \cite{levitan2012acoustic, litman2016teams}, lexical \cite{brennan1996lexical}, and syntactical \cite{branigan2000syntactic}. It has long been an interest of dialogue studies because entrainment has been found to associate with various social outcomes such as group relationship \cite{yu2019investigating}, positive or negative effect \cite{nasir2019modeling}, being liked by partners \cite{levitan2012acoustic}, and dialogue success \cite{kawano2020entrainable,xu2017spectral}. The characteristics make entrainment a valuable tool to build a human-like dialogue system.

Here we are specifically interested in studying entrainment based on text features. One popular type of study focuses on lexical entrainment. While there are various approaches to quantify lexical entrainment, the core instrument of those approaches is a linguistic similarity measure between conversational partners \cite{brennan1996conceptual, brennan1996lexical, ward2007automatically, nenkova2008high, rahimi2017entrainment, van2009lexical, stoyanchev2009lexical}. Most current approaches are built upon bag-of-words models that rely heavily on linguistic markers such as function words or high-frequency words \cite{rahimi2017entrainment, nenkova2008high, gonzales2010language, yu2019investigating, pennebaker2007liwc2007}. However, linguistic markers are insufficient to capture context, irony, sarcasm, or other word semantics \cite{pennebaker1999linguistic}. Sparsity caused by low-level word usage raises reliability concern for this type of measure \cite{zeldow1993comparison}. While more advanced measures in recent entrainment studies are starting to utilize word representation enriched with semantics such as word embeddings \cite{nasir2019modeling}, the primary comparison granularity is still single words isolated from the conversation flow.

Therefore we propose an alternate approach using neural networks to perform similarity measures for entrainment calculation. Neural network models are data-driven and are highly self-governing. Using neural network-based models allows us to decouple the entrainment similarity measure from the bag-of-words paradigm. Specifically, input sequences can be represented by high-dimensional vectors embedded with semantic meaning. Beyond word-level information, using sequential architectures such as Long-Short Term Memory Network (LSTM) \cite{greff2016lstm}, the model can learn structural dependencies among input units at different levels. Feature extraction is also fully automated in neural-based models. 




Beyond using a neural network framework, we attempt to learn the high-level linguistic features beyond the inherent text representation of dialogues. The conventional similarity measure for entrainment often leverages high-level linguistic features that can be shared across conversations, such as corpus topics \cite{rahimi2017entrainment}, high-frequency words \cite{nenkova2008high}, and general language style reflected by function words \cite{gonzales2010language}. To simulate this mechanism, we introduce an attention-based architecture to our neural model to generalize high-level linguistic features shared across all input dialogues. These high-level features are supposed to be global and agnostic to the actual content and input forms, leading to a better representation generalization for unstructured data. The architecture is inspired by Global Style Token (GST) that have been previously used in the speech synthesis to generalize speech styles \cite{wang2018style, an2019learning}. Similar architectures have been adopted in other research area such as machine translation task \cite{wang2019multilingual} and dialogue response matching \cite{humeau2019poly}. We don't limit our high-level linguistic features to any specific types. Instead, due to the nature of neural network features learning, the high-level features can describe a comprehensive set of features such as language style, sentence structure, and semantics.
We name this component \textit{shared stylebook} as their parameters are globally shared across all inputs. The ``style'' in our shared stylebook has a broader definition. 

Our ultimate goal is to improve the comprehensiveness of lexical entrainment by using our model to perform the similarity measure. The ``lexical entrainment'' we refer to in this study has a broader definition beyond lexical.

In this study, we examine 2 specific hypotheses:
\textbf{Hypothesis 1}: Leveraging high-level features will aid input representation, leading to a more robust model.  
\textbf{Hypothesis 2}: Our neural network-based measures will capture a stronger entrainment signal compared to the bag-of-words measures.
The results show that both of our hypotheses are positive.

\section{Related Work}
\label{sec:related_work}
\textbf{Matching Dialogue Response Selection} Our model follows neural dialogue response matching frameworks. Matching between the dialogue context and responses is a trendy task in building retrieval-based dialogue systems. The neural network-based models received the most attention in recent years. 
Early studies focus on single-turn interactions that only considers the dialogue context as a single query by concatenating all previous turns \cite{yan2016learning, lowe2015ubuntu, wang2016machine}. Later studies are more interested in learning multi-turn interactions so that the multiple turns in the context are all used as separate queries \cite{zhou2018multi, lu2019constructing, tao2019one}. Recent studies show increasing interests in using pre-trained language models such as BERT \cite{devlin2018bert, wu2020enhancing, bertero2020Model}. Our work focuses on building a single-turn dialogue response matching model. \textit{Compared to the existing single-turn model, we add a component to facilitate learning high-level linguistic features.}

\noindent\textbf{Style Response Generation/Selection} Because we attempt to study the ``style'' of language, another closely related research topic is dialogue style generation or selection. 
One typical strategy to generate stylized dialogue responses is to employ 2 separate training stages for response generation and style controlling. Works in style controlling use different approaches such as pre-training stylized language models \cite{niu2018polite}, fine-tuning model with styled corpus \cite{akama2017generating}, using adversarial training \cite{zheng2020stylized}, and learning a shared latent space between a response and stylized sentences \cite{gao2019structuring}. Generating personalized \cite{li2016persona} or emotional responses \cite{zhou2017emotional} are also in the same category since they all require controlling some type of style. Our study specifically focuses on dialogue style matching, which has been viewed as a subtask in some style generation models \cite{luo2018auto, niu2018polite}. \textit{Compared to previous studies, rather than a well-defined style, the style learned by our model is generalizing from the input corpus.}

\noindent\textbf{Linguistic Entrainment}
There has been substantial evidence for entrainment in many linguistic dimensions, such as acoustic-prosodic entrainment \cite{levitan2011measuring, levitan2012acoustic, ward2007automatically}, lexical \cite{brennan1996conceptual, brennan1996lexical, ward2007automatically}, and syntactic entrainment \cite{branigan2000syntactic, stoyanchev2009lexical, cleland2003use}. 
To evaluate entrainment, early studies often set experimental conditions or control groups \cite{brennan1996conceptual, branigan2000syntactic, garrod1987saying}. In the later corpus-based studies, evaluations are mostly \textbf{extrinsic} such as comparing entrainment between conversational partners and non-partners \cite{levitan2011measuring, rahimi2017entrainment}, associating entrainment to other interpersonal behaviors in dialogue such as group relationships \cite{yu2019investigating}, positive or negative effects \cite{nasir2019modeling}, being liked by partners \cite{levitan2012acoustic}, and dialogue success \cite{kawano2020entrainable}. \textit{Here, we evaluate our entrainment measures by predicting dialogue success reflected by social outcomes.}

\section{Data}
\label{sec:data}
In this study, we will focus on a constrained non-goal oriented dataset.
The Teams Corpus is a small-scale multiparty spoken dialogue dataset \footnote{https://sites.google.com/site/teamentrainmentstudy/corpus} \cite{litman2016teams}. It consists of 124 multiparty conversations (62 for Game 1 and 62 for Game 2) elicited from 213 native speakers of American English. Each group of speakers participated in a collaborative game called Forbidden Island. This dataset provides transcriptions and surveys evaluating speaker personality and group relationships. Humans transcribe each inter-pausal unit (IPU). Here we will view each IPU as a conversational turn. We choose this dataset because there are many entrainment-related studies on this dataset using the bag-of-word paradigm to establish a benchmark for entrainment analysis. 


\section{Model}
Our model is a neural dialogue response matching model. It measures the matching between a dialogue context and a response, and it can be used as a similarity measure for entrainment. We train and evaluate the model following the standard framework of dialogue response matching task defined in the following Section \ref{sec:problem_formulization}. 

\subsection{Problem Formalization}
\label{sec:problem_formulization}
\begin{figure}[htbp]
\centering
\includegraphics[width=1\columnwidth]{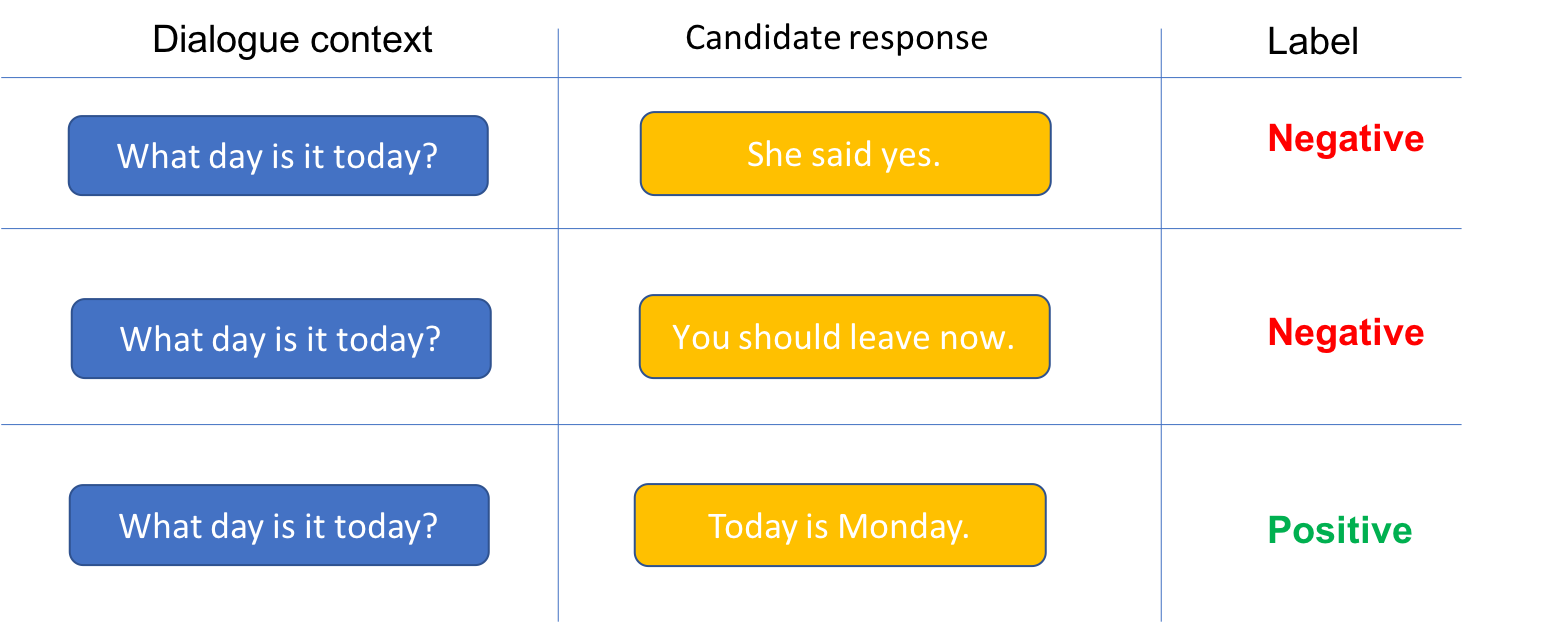}
\caption{Input examples for the dialogue response matching task.}
\label{fig:input_examples}
\end{figure}

Given a dialogue context, response matching models determine whether an utterance is proper as a response. Formally, each train and test example is a triplet (\textbf{C}, \text{R}, y) where \textbf{C} is the dialogue context, \text{R} is a response, and y is a label indicating whether \text{R} is proper for \textbf{C}. Given a dialogue \textbf{D}
$={u_1, u_2,...,u_n}$ where $u_i$ is the utterance for i-th turn, we can extract a dialogue context \textbf{C} = ${u_1, u_2,...,u_{n-1}}$, a ground truth response \textbf{R} = ${u_n}$, and we can randomly sample false responses {R$^\prime$} from the same corpus. Therefore, we can formulate our task as a binary classification task to determine $y\in(0,1)$ for each (\textbf{C}, \text{R}, y) as $y=1$ indicating the ground truth. A candidate response is positive when $y=1$ and negative when $y=0$. Figure \ref{fig:input_examples} shows 3 input examples.

\subsection{Model Design}
\label{sec:NLM_training}
\begin{figure*}[htbp]
\centering
\includegraphics[width=1.4\columnwidth]{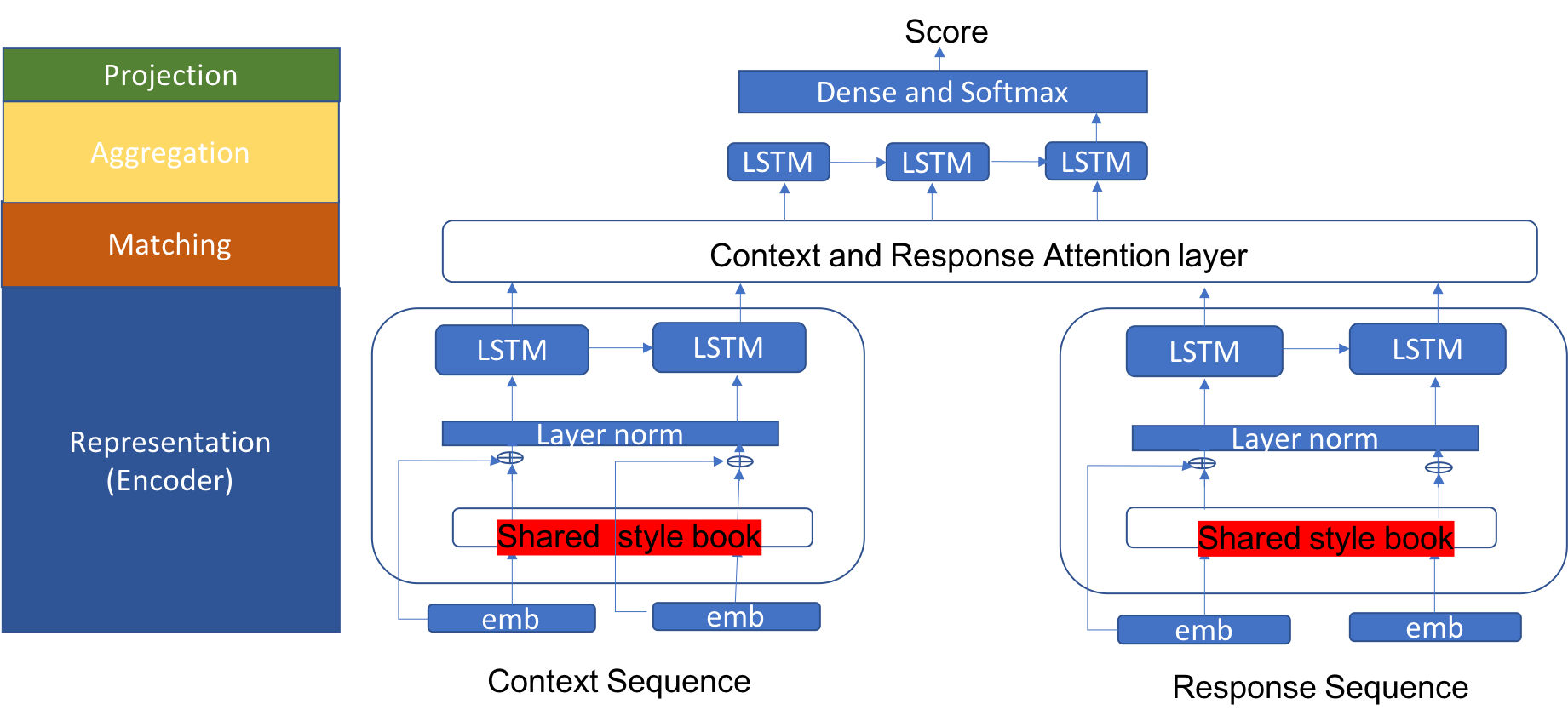}
\caption{Model illustration. The part highlighted with red is the stylebook.}
\label{fig:matching_model}
\end{figure*}

Following the practice in prior works \cite{zhou2018multi, yan2016learning, wu2020enhancing, bertero2020Model, lu2019constructing, tao2019one}, we train our model with a binary classification objective. We adopted a representation-matching-aggregation framework used in previous works \cite{zhou2018multi, wu2016sequential}. 
Figure \ref{fig:matching_model} is the model illustration. Note that state-of-the-art dialogue response matching models are mostly multi-turn models. 
Our model is single-turn because multi-turn models substantially benefit from learning turn interactions by complicated models. To avoid that and focus on our goal in this study, we choose to follow a simpler single-turn model design.

\subsubsection{Representation (Encoder)}
\label{sec:reranker_enc}

\noindent \textbf{Embedding Layer}
The embedding layer transforms our input of subword tokens to high-dimensional continuous representations. Given a dialogue context \textbf{C} and a response candidate \textbf{R}, the representations are  \textbf{C} = [$e_{c_{w_{1}}}$,...$e_{c_{w_{n}}}$] and \textbf{R} = [$e_{r_{w_{1}}}$,...,$e_{r_{w_{n}}}$], where $e_{c_{w_{i}}}$ and $e_{r_ {w_{i}}}$ represents the embeddings of the i-th token of \textbf{C} and \textbf{R} respectively. Here \textbf{C} $\in \mathbb{R}^{ {n_c} \times{d}}$ and \textbf{R} $ \in \mathbb{R}^{{n_r} \times{d}}$ where ${n_c}$, ${n_r}$ and d denotes the number of tokens in the context, the number of tokens in the response, and the embedding size, respectively.  \\

\noindent \textbf{Shared StyleBook} 
\label{sec:shared_style_book}
The stylebook consists of a set of randomly initialized global key-value pairs. \textit{Unlike the self-attention \cite{vaswani2017attention} that the key (K) and value (V) are the linear transformation of input query (Q) itself, our K and V are global for all Q.} The stylebook is followed by a multi-head scaled dot-product attention \cite{vaswani2017attention} that performs as a similarity metric between the key-value set and the input embeddings.
Equations \ref{eq:singlehead_attn} and \ref{eq:multihead_attn} define the attention function where the query (Q) is the input embeddings of the encoders. Specifically, Q is equivalent to \textbf{C} in the context encoder or \textbf{R} in the response encoder. 
V denotes value consisted of randomly initialized weights so that $V \in \mathbb{R}^{{T} \times{d_v}}$ where ${T}$ and $d$ denote the size of the stylebook and its dimension. 
Here we let $d$ be the same as the embedding dimension because we will apply a residual connection later. Key (K) is a linear transformation of V. 

\begin{equation}
\centering
    \text{Attention}(Q,K,V) = \text{softmax}(\frac{QK^T}{\sqrt{d_k}})V
\label{eq:singlehead_attn}
\end{equation}

\begin{equation}
\centering
\begin{split}
    \text{MultiHead}(Q,K,V) &= \text{Concat}(head_1, .., head_n)\\
\text{where }head_i &= \text{Attention}(Q_i, K_i, V_i)
\end{split}
\label{eq:multihead_attn}
\end{equation}

\noindent For each head \textit{i} in \textit{n} heads, we have $Q_i$, $K_i$, $V_i$ that $Q_i \in \mathbb{R}^{{n_q}\times {d_{i}}} $, $K_i \in \mathbb{R}^{T \times{d_{i}}} $,$V_i \in \mathbb{R}^{T \times{d_{i}}}$, where ${n_q}$, T and $d_{i}$ are the query length, the size of the stylebook and the size of each head. The output of the attention layer is a similarity matrix $M_{\text{style}} \in \mathbb{R}^ {n_q \times d}$ where $n_q = n_c$ for context and $n_q = n_r$ for the response. We can view this similarity matrix as style embeddings for they represent the contribution of input embeddings on each type of ``style'' in the stylebook.
 We employ a residual connection and layer normalization (Add\&Norm) after the attention. Thus the final output is a hybrid embedding vector that combines the inherent and style embeddings, which is denoted as $C_{hybrid} \in \mathbb{R}^ {n_c \times d}$ for the context and $R_{hybrid} \in \mathbb{R}^ {n_r \times d}$ for the response. \\

\textbf{LSTM Layer}
We choose to use an LSTM to learn the dependencies and temporal relationships between input features. LSTMs are a popular type of RNN to model sequential inputs for its prominent ability to control the short-term or long-term information. 
In our case, the inputs for this layer are hybrid embeddings from the stylebook, and the outputs are hidden states for each time step denoted by $H_c \in \mathbb{R}^ {n_c \times d_h}$ for \textbf{C} , and $H_r \in \mathbb{R}^ {n_r \times d_h}$ for \textbf{R}, where $d_h$ is the number of hidden units. We will use $H_c$ and $H_r$ as the final context and response encodings generated from the encoders. 

\subsubsection{Matching}
This layer performs the matching between context and response encodings. We use the scaled dot-product attention \cite{vaswani2017attention} to measure the similarity between context encodings $H_c$ and response encoding $H_r$.
Specifically the query Q is the response encoding $H_r$, and the key-value pairs are from context encoding $H_c$. This allows a response to query the most related context information stored in value. Thus, each element in the resulting matrix reflects the similarity between the response and context until the i-th text segment. 
The layer output is a similarity matrix $M_{r,c}$, which $M_{r,c} \in \mathbb{R}^{n_r \times d}$.



\subsubsection{Aggregation}
Similar to prior works in neural matching networks \cite{wu2016sequential, zhou2018multi, lu2019constructing}, we use an aggregation layer to aggregate matching across segments. Our model aggregates all the segmental matching given by $M_{r,c}$ using an LSTM layer. We use the last hidden state $h_{n_r}$ from the aggregation layer as the sequence-level matching. 

\subsubsection{Projection}
\label{sec:projection}
The output vector $h_{n_r}$ will be fed into a dense layer followed by a softmax layer. The output probability is used as the matching score g between the context \textbf{C} and a response candidate \textbf{R}. Formally, the g is calculated as in Equation \ref{eq:score}. \textit{The matching score g will be used as the similarity measure in the entrainment task.}

\begin{equation}
\centering
g(C,R) = \text{softmax}(\textbf{W} h_{n_r} + \textbf{b})
\label{eq:score}
\end{equation}

where \textbf{W} and \textbf{b} are learned parameters.

\section{Measuring Entrainment}


\subsection{Train the Matching Model}
We firstly train our matching model on the dataset. 
We create a Teams Corpus dataset for dialogue response matching task (see Section \ref{sec:problem_formulization}). 
We sampled examples from each dialogue. To make an example, we extract the previous 5 turns as the dialogue context and the following turn as the ground truth responses. This process results in 107,420 positive instances. We split positive instances in train, validation, and test based on a ratio of 6:2:2. Then for each positive instance, we randomly sampled 9 false responses for validation and test sets, and 1 false response for the train set. This operation results in a dataset of 129K, 215K, 215K examples in train, validation, and test set.


\subsection{Measuring Entrainment as Convergence}
\label{sec:measuring_entrainment}
Our approach to measure entrainment in Teams Corpus is based on \citet{yu2019investigating}. For each conversation in the corpus, we first split it into 10 equivalent time intervals. For an utterance \textit{i} in the interval \textit{j}, we use above model to score the similarity between \textit{i} and the dialogue context \textbf{C} consisted of the previous 10 turns. Equation \ref{eq:speaker_score} shows the calculation. \textit{g(C, i)} denotes the model generated matching score between context \textbf{C} and \textit{i} (see Section \ref{sec:projection}). Then we average the similarity score over the total \textit{n} utterances spoken by a speaker during interval \textit{j}.  \citet{yu2019investigating} use a bag-of-words based similarity score to quantify group difference, and then calculate convergence. Note that the baseline has 2 types of bag-of-words similarity scores depending on different algorithms, but we do not worry about them here because we will replace the bag-of-words score with our neural one. 
Defined in Equation \ref{eq:TDiff_speaker}, team difference (\textit{TDiff}) is the averaged similarity difference for pair-wise speakers supposing there are \textit{m} speakers speak in the interval \textit{j}. Shown in Equation \ref{eq:dialogue_convergence}, entrainment is measured as the convergence, which indicates the increase in partner similarity, between 2 arbitrary intervals \textit{q} and \textit{p} with \textit{q} being earlier than \textit{p}. To obviate the need to select time intervals, 4 types of convergence variables are derived from ${C}_{pq}$: \textit{Max}, \textit{Min}, \textit{absMax}, \textit{absMin}. The calculation formulas of \textit{Max} and \textit{absMax} are shown in Equation \ref{eq:convergence_Max}. \textit{Min} and \textit{absMin} are calculated similarly. To summarize, \textit{compared to \citet{yu2019investigating}, we use their group difference and convergence formula, but with a more advanced model-generated similarity score}.

\begin{equation}
{Score}_{speaker} =  \frac{\sum^{n}_{i}{g(C, i)}}{n}
\label{eq:speaker_score}
\end{equation}

\begin{equation}
{TDiff}_{j} = \frac{\sum_{\forall a, b \in m}(|{Score}_a-{Score}_b|)}{|m| * (|m| - 1)}
\label{eq:TDiff_speaker}
\end{equation}

\begin{equation}
C_{qp} = TDiff_q-TDiff_p, q < p <= 10
\label{eq:dialogue_convergence}
\end{equation}

\begin{equation}
Max = Max\{C_{ij} > 0 \},  absMax = Max \{|C_{ij}|\}
\label{eq:convergence_Max}
\end{equation}

\section{Experiments}
\label{sec:experiments}

\subsection{Hypothesis 1 (H1)}
We first hypothesize that leveraging high-level features will aid input representation, leading to a more robust model in matching dialogue responses. We train 2 models on Teams Corpus to examine this hypothesis: one is our proposed model, and another one is a baseline model with the stylebook removed. Next, we determine whether our model outperforms the baseline model.

\subsubsection{Evaluation Metrics}
\label{sec:LSM_evalmetrics}
We follow the standard metrics of dialogue response matching task to evaluate Recall@1 (R@1), Recall@2 (R@2), and Recall@5 (R@5). The $k$ in the Recall@k means that the true positive response is among the first $k$ ranked candidates. 


\subsubsection{Implementation Details}
\label{sec:eval_details}
Our model is implemented in Pytorch and trained using 3 GPUs. We use pre-trained English byte pair embeddings (bpemb) from BPEmb \cite{heinzerling2017bpemb}. 
Model configuration is tuned on the validation set. The embedding dimension is 300. The maximum token length is 40 for the context and 20 for the response.
The size of the stylebook is set to 500. Encoders are shared between the context and the responses. The LSTM layer in encoders has 1024 hidden units. The aggregation LSTM layer has 128 hidden units. All multi-head attention used in this model have 4 heads. Models are trained in mini-batches with a size of 128. The learning rate is 0.0001. We use Adam optimizer. The loss function is cross-entropy. We train a maximum of 10 epochs and optimize training at the R@1 on the validation set.

\subsubsection{Model Performance}
\label{sec:model_performance}

Table \ref{tab:model_baseline_comparison} shows the evaluation results. Our proposed model with the stylebook overall outperforms the baseline. Without the stylebook, the model performance decays a margin. The R@1, R@2, R@5 decrease 3.7\%, 5.1\%, and 3.5\%, respectively. 
We also compare the number of model parameters. We observe only a minimal growth of parameter size. Beyond Teams data, we further test our stylebook model on another 2 dialogue response matching datasets and similarly observe an  improvement in the  model performance. Further experiment details are given in the Appendix.

\begin{table}[h]
\centering
\resizebox{0.85\columnwidth}{!}{%
\begin{tabular}{|c|cccc|}
\hline
 & R@1 & R@2 & R@5 & Size \\ \hline
Our model              & \textbf{24.9\%}                   & \textbf{41.8\%}                  & \textbf{74.7\%}  & 12.4M\\ 
- stylebook       & 21.2\%                   & 36.7\%                   & 71.2\% & 12.2M \\ \hline 
\end{tabular}
}
\caption{Model performance on Teams Corpus for the dialogue response matching task. Size: model size}
\label{tab:model_baseline_comparison}
\end{table}

\subsubsection{Understanding the Stylebook}
\begin{figure*}
        \begin{subfigure}[b]{0.323\textwidth}
                \centering
               \includegraphics[width=\linewidth]{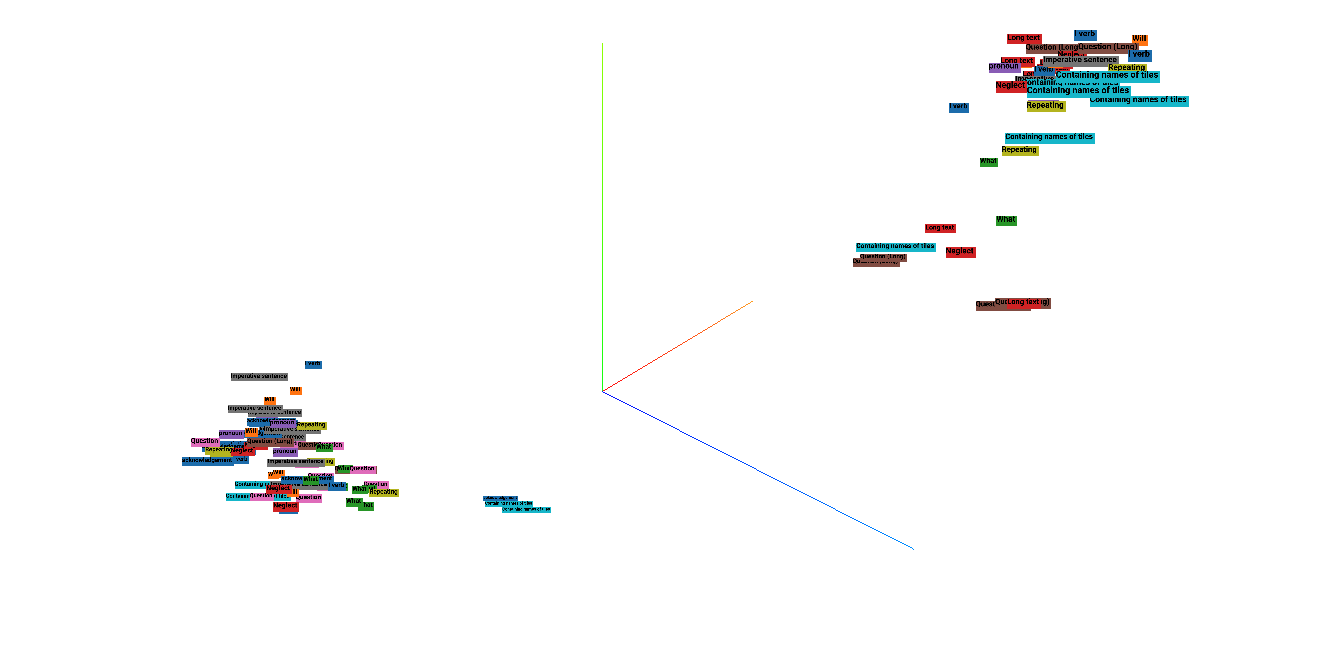}
                \caption{Clustering overview shows 2 major clusters.}
                \label{fig:overview}
        \end{subfigure}%
        \hfill
        \begin{subfigure}[b]{0.323\textwidth}
                \centering
               \includegraphics[width=\linewidth]{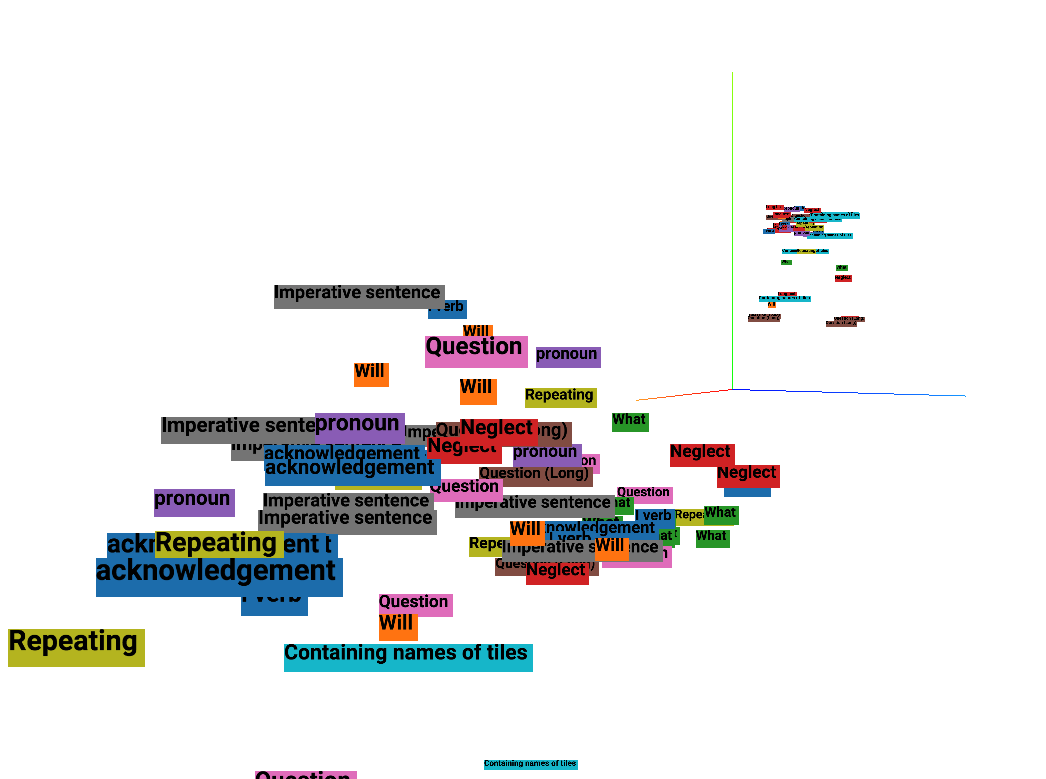}
                \caption{A focus view of Cluster 1, which contains many short sequences.}
                \label{fig:cluster1}
        \end{subfigure}%
        \hfill
        \begin{subfigure}[b]{0.323\textwidth}
                \centering
                \includegraphics[width=\linewidth]{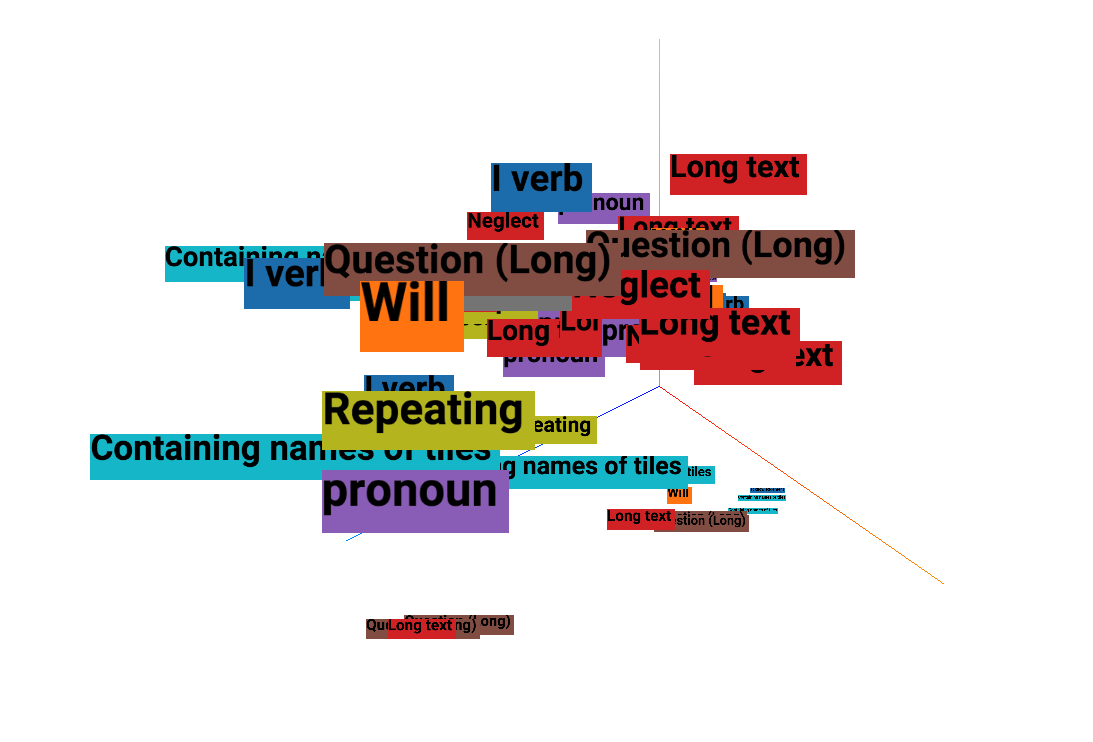}
                \caption{A focus view of Cluster 2, which contains many long sequences.}
                \label{fig:cluster2}
        \end{subfigure}%
        
\caption{3D T-SNE visualization of style embeddings. Each data point represents an utterance displayed by its category in a distinct background color. }
\label{fig:tsne}
\end{figure*}

\textbf{Visualization}
To understand whether the stylebook learns any style as we expect, we conduct a study to visualize the style embeddings (see Section \ref{sec:shared_style_book}) generated from the stylebook. 
We hand-label 130 utterances from the Teams Corpus with 13 categories of styles based on our intuition. For example, utterances claiming acknowledgement such as ``yeah'' and ``yes'' are categorized as \textit{Acknowledgement}. We collect 10 utterances for each category. Please see the appendix for the description of the 13 categories. We extract the averaged style embeddings of these sequences \footnote{The style embeddings are extracted from a smaller model with 300 LSTM hidden units.} and then project them into a 3D space using t-distributed stochastic neighbor embedding (t-sne) \cite{van2008visualizing} with a perplexity of 5 and learning rate of 1 \footnote{We use an embedding projector provided in \citet{smilkov2016embedding}.}. Figure \ref{fig:tsne} shows the results. Each data point in the figure represents an utterance, and it is displayed in the figure by its category in a distinct background color. Figure \ref{fig:overview} is a clustering overview that shows 2 major clusters. Figure \ref{fig:cluster1}  and \ref{fig:cluster2} are the focus views of the first (Cluster 1)  and the second cluster (Cluster 1) respectively. Cluster 1 mainly contains short utterances such as questions, acknowledgments, and questions starting with ``What''. As a counterpart, Cluster 2 mainly contains long utterances such as long questions consisting of more than 2 short questions. In each cluster, utterances belonging to the same category are more likely to be located closely, indicating that the style embeddings can identify utterances with similar characteristics. 

\subsection{Hypothesis 2 (H2)}
We hypothesize that our neural network-based measures will capture a stronger entrainment signal compared to the bag-of-words measures. A recent study \cite{rahimi2020entrainment2vec} on Teams Corpus suggests that more robust entrainment measures carrying stronger signals will lead to a more robust prediction of dialogue outcomes. Thus, we examine this hypothesis with an \textbf{extrinsic} evaluation to predict dialogue success on Teams Corpus.

\subsubsection{Baseline Models}
The baseline is a bag-of-word approach from \citet{yu2019investigating} on the Teams Corpus. We only focus on Game 1 as the baseline performance was only published for that portion of the Teams data.

\subsubsection{Validate our Measure}
Before the prediction, we first validate our similarity measures to ensure they convey some linguistic signals associated with entrainment. Thus we calculate the Pearson correlations between our baseline convergence variables and their baseline counterparts. Note that the baseline approach provides 2 types of convergence variables of being weighted and unweighted based on different bag-of-words algorithms. Furthermore, to investigate the impact of the stylebook in our model, we remove the stylebook and examine the correlations again. The results are shown in  Table \ref{tab:correlation_convergence}. We first find that our \textit{Max} and \textit{absMax} are strongly correlated to the baseline convergence variables. This finding suggests that the neural model can be used as a similarity measure for entrainment due to its correlation. On the other hand, the correlation becomes much weaker if we eliminate the stylebook from our model. Intuitively, this finding implies that the stylebook may contribute to capture the linguistic signal related to entrainment.


\begin{table}[]
\centering
\resizebox{\columnwidth}{!}{%
\begin{tabular}{|l|c|ccc|ccc|}
\hline
&            &  \multicolumn{6}{c|}{Baseline}               \\\hline 
&            &  \multicolumn{3}{c|}{Weighted}              &  \multicolumn{3}{c|}{Unweighted}               \\ 
&            & absMax             & Max &Min  & absMax           & Max  &Min    \\ \hline
Ours&absMax & 0.426** & - &- & 0.316* & - &-\\ 
&Max   & -       & 0.419**  & - &- & 0.351** & - \\ 
&Min   & -       & -  & Not sig & - & - & .290*\\ \hline
-stylebook&absMax & 0.257* &- &-& Not sig & - &- \\ \hline
\end{tabular}}
\caption{The Pearson correlations between our baseline convergence variables and their baseline counterparts. Rows only show our variables that have at least one significant correlation. Not sig: not significant. * if p$<$0.05, ** if p$<$0.01}
\label{tab:correlation_convergence}
\end{table}

\subsubsection{Evaluation Method}
\label{sec:H2_eval_method}
We follow the baseline approach to evaluate entrainment measures by predicting dialogue success using a regression model. 
Formally, entrainment measures are used as the independent variables (IVs) to predict dialogue success measures as the dependent variables (DVs). The DVs are entered into the model stepwise. We construct both IVs and DVs strictly following the baseline. DVs are 4 social outcome scales extracted from Teams Corpus surveys: \textbf{Team Processes}, \textbf{Task Conflict}, \textbf{Process Conflict} and \textbf{Relationship Conflict}. \textbf{Team Processes} is an aggregated scale of team cohesion, general team satisfaction, potency/efficacy, and perceptions of shared cognition \cite{wendt2009leadership,wageman2005team,guzzo1993potency,gevers2006meeting}. Conflict scales reflect the conflicts in completing tasks, work processes, and interpersonal relationships. IVs are convergence variables described in the Equation \ref{eq:convergence_Max}. 


\subsubsection{Predicting Dialogue Success}

Table \ref{tab:regression} shows standardized coefficients ($\beta$), $R^2$ and F value of a regression model with statistical significance. Here we construct 3 models: \textbf{Baseline} is the baseline model that predicts DVs by bag-of-words entrainment measures. 
The result of \textbf{Baseline} is copied directly from the previous work. \textbf{Ours} predicts DVs by our neural network-based entrainment measures. Additionally, \textbf{No Stylebook} predicts DVs by our neural network-based entrainment measures with no stylebook removed from the model structure. Results show that overall both our neural model \textbf{Our} and \textbf{No Stylebook} are stronger in predicting all DVs reflecting all \textbf{Conflicts} variables. For explaining variation in \textbf{Task Conflict}, \textbf{Baseline} only archives significant $R^2$ of 7\%, but using entrainment measures from \textbf{Ours} and \textbf{No Stylebook}, the resulted $R^2$ is highly significant. \textbf{No Stylebook} achieves the highest $R^2$ improvement of 7\% compared to the \textbf{Baseline}. We have the same finding for \textbf{Process Conflicts}. Although the improvement in $R^2$ between \textbf{Baseline} and our models are smaller, $R^2$ of \textbf{Ours} and \textbf{No Stylebook} are highly significant. More notably, using our neural entrainment measures, we can predict \textbf{Relationship Conflict}, which was not predictable by the \textbf{baseline}. Both \textbf{Our} and \textbf{No Stylebook} achieve significant 8.0\% and 11\% $R^2$ for \textbf{Relationship Conflict}. \textbf{No Stylebook} is a highly significant regression model. Therefore, we conclude that our neural-based entrainment measures are stronger in predicting all DVs reflecting Conflicts compared to the baseline model. Beyond the performance improvement, we found that \textbf{No Stylebook} performed better than \textbf{Our} having the stylebook in its model structure. This implying that the improvement in regression was not caused by using the stylebook. 
We also noticed that the most predictive IV across all 3 models is \textit{absMax}, which represents the maximum magnitude of convergence. Also, negative entrainment coefficients reveal that a higher convergence signals less conflict in the conversation. This finding is aligned with existing findings.

\begin{table}[h]
\centering
\resizebox{1\columnwidth}{!}{
\begin{tabular}{cllccc}
\hline
\textbf{DV}      &       & \textbf{IV}         & \textbf{Baseline ($\beta$)} & \textbf{Our($\beta$)} & \textbf{No Stylebook($\beta$)}  \\
\hline

Task Conflict        &       & Baseline ent (\textit{unw. absMax})     &  -0.27*   & -      & -  \\
        &       & Our ent (\textit{absMax})  &  -        & -0.35** &   -        \\
        &       & No Stylebook ent (\textit{absMax})  &  -        & - &   -0.37**        \\

        \cline{3-6}
        & $R^2$ &                   &   0.07     & 0.12    &    \textbf{0.14}      \\
        & F     &                   &   4.77*    & 8.10**            &  9.73** \\
        
\hline
Process Conflict        &       & Baseline ent (\textit{w. absMax})         & -0.34**    &   - & -  \\
        &       & Our ent (\textit{Max})         &  -         & -0.34** & - \\
          &       & No Stylebook ent (\textit{absMax})         &  -         & - & -0.37**\\
        \cline{3-6}
        & $R^2$   &        & 0.12      &   0.12    &    \textbf{0.14}   \\
        & F        &        & 7.87**    &   7.85**          & 9.44** \\
\hline
Relationship Conflict         &       & Baseline ent         & -    &   - & -  \\

       &       & Our ent (\textit{absMax})    & -      & -0.28* & -\\
        &       & No Stylebook ent (\textit{absMax})    & -      & - & -0.33**\\
        \cline{3-6}
        & $R^2$ &               & -      & 0.08    &   \textbf{0.11}    \\
        & F     &                & -      & 5.20*       &  7.21**   \\
\hline
\end{tabular}}
\caption{3 regression models. Baseline ent, Our ent, No Stylebook ent denote the entrainment convergence variables derived by the baseline approach, our neural model, our neural model after removing the stylebook, respectively. w. absMax:weighted absMax, unw. absMax:unweighted absMax, * if p $<$0.05, ** if p $<$0.01.}
\label{tab:regression}
\end{table}

\begin{figure*}[ht]
\centering
\includegraphics[width=1.9\columnwidth]{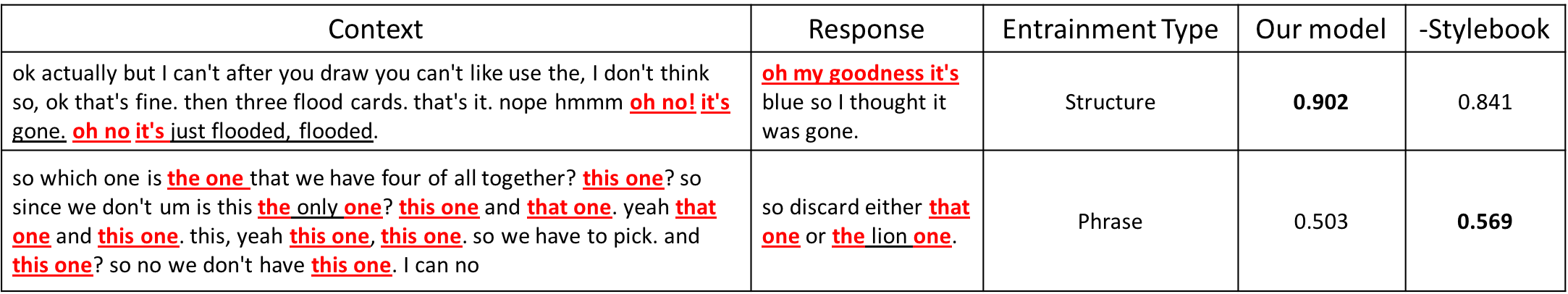}
\caption{2 dialogue examples. Previous 10 turns (IPUs) are concatenated as the context. Our model and -Stylebook show the similarity scores from our model before and after removing the stylebook. }
\label{fig:case_study}
\end{figure*}

\section{Case Study}
We perform a case study on the similarity score \textit{g} (see the Equation \ref{eq:score}) generated by our model to investigate how it reflects entrainment
and whether the shared stylebook has any impact on that. Based on our interpretation, we cherry-pick 2 Teams Corpus dialogue examples that exhibit different types of entrainment. Table \ref{fig:case_study} shows the dialogue context, the ground truth response, and the scores generated from our neural model when using and not using the shared stylebook. 
In the first example, similarly to the context, the response contains an exclamation immediately following a phrase starting with  ``it's''. This example can be interpreted as a case of structure entrainment because the same sentence structure is repeated in the response. In the second example, the phrases ``this one'' and ``that one'' are frequently used in the context. The response also contains ``that one'', and more notably, the speaker chooses to say ``the lion one'' when there is a simpler alternation ``the lion''. We interpret this example as a phrase entrainment. For the model-generated similarity scores, the neural models, including the model using and not using the stylebook, score more than 0.5 out of 1, indicating that the models likely predict the response as a proper response for the context. Although we expect using the stylebook would aid the model in recognizing entrained cases, the model using the stylebook does not assign higher scores for both of the entrained cases than the model not using the stylebook. This observation is not surprising because neural models, in general, are difficult to interpret. 
The high-level linguistic features may not align with our intuition.

\section{Conclusion and Future Work}
We present a neural dialogue response matching model specifically designed as a similarity measure for lexical entrainment. We propose to leverage the shared stylebook to generalize the high-level shared linguistic features across dialogues. The results suggest that the shared stylebook improves the model performance in a dialogue response matching task. 
We perform several ablation studies to understand the impact of the stylebook and the underlying meaning of its embeddings. 
We find our similarity measure is strongly correlated with an existing bag-of-words entrainment measures. 
Removing the stylebook will weaken the correlation, implying that the stylebook is vital for generating meaningful entrainment measures.
We then conduct an extrinsic evaluation to compare our measure and the bag-of-words measure in dialogue outcome prediction. Our measure leads to a more robust prediction model with a stronger entrainment signal. On the other hand, the improvement of prediction is not caused by using the stylebook. The Extrinsic evaluation has its limitation. 

\section{Acknowledgement}
This study is inspired and partially based upon an unpublished work of Mingzhi Yu during her internship in Microsoft Corporation, Redmond. We thank Microsoft researchers for their valuable feedback. 

\bibliographystyle{acl_natbib}
\bibliography{emnlp2021}

\end{document}